\documentclass[10pt,twocolumn,letterpaper]{article}

\usepackage[pagenumbers]{cvpr} 

\usepackage[dvipsnames]{xcolor}

\definecolor{cvprblue}{rgb}{0.21,0.49,0.74}
\usepackage[pagebackref,breaklinks,colorlinks,citecolor=cvprblue]{hyperref}

\usepackage{graphicx}
\usepackage{amsmath}
\usepackage{amssymb}
\usepackage{booktabs}
\usepackage{multirow}
\usepackage{makecell}
\usepackage[pagebackref,breaklinks,colorlinks]{hyperref}
\usepackage{bm}
\usepackage{enumitem}
\usepackage[capitalize]{cleveref}

\def\paperID{8465} 
\def\confName{CVPR}
\def\confYear{2024}

\title{Locally Adaptive Neural 3D Morphable Models}

\author{Michail Tarasiou \qquad Rolandos Alexandros Potamias \qquad Eimear O'Sullivan \\ 
Stylianos Ploumpis \qquad Stefanos Zafeiriou \\
Imperial College London \\
{\tt\small \{michail.tarasiou10,r.potamias19,e.o-sullivan,s.ploumpis,s.zafeiriou\}@imperial.ac.uk}
}

\begin{document}
\maketitle
\begin{abstract}
We present the Locally Adaptive Morphable Model (LAMM), a highly flexible Auto-Encoder (AE) framework for learning to generate and manipulate 3D meshes. 
We train our architecture following a simple self-supervised training scheme in which input displacements over a set of sparse control vertices are used to overwrite the encoded geometry in order to transform one training sample into another. 
During inference, our model produces a dense output that adheres locally to the specified sparse geometry while maintaining the overall appearance of the encoded object. This approach results in state-of-the-art performance in both disentangling manipulated geometry and 3D mesh reconstruction. To the best of our knowledge LAMM is the first end-to-end framework that enables direct local control of 3D vertex geometry in a single forward pass. 
A very efficient computational graph allows our network to train with only a fraction of the memory required by previous methods and run faster during inference, generating 12k vertex meshes at $>$60fps on a single CPU thread. 
We further leverage local geometry control as a primitive for higher level editing operations and present a set of derivative capabilities such as swapping and sampling object parts. Code and pretrained models can be found at \verb!https://github.com/michaeltrs/LAMM!.
\end{abstract}    
\section{Introduction}
\label{sec:intro}

\begin{figure}[t]
    \centering 
    \includegraphics[width=0.4\textwidth]{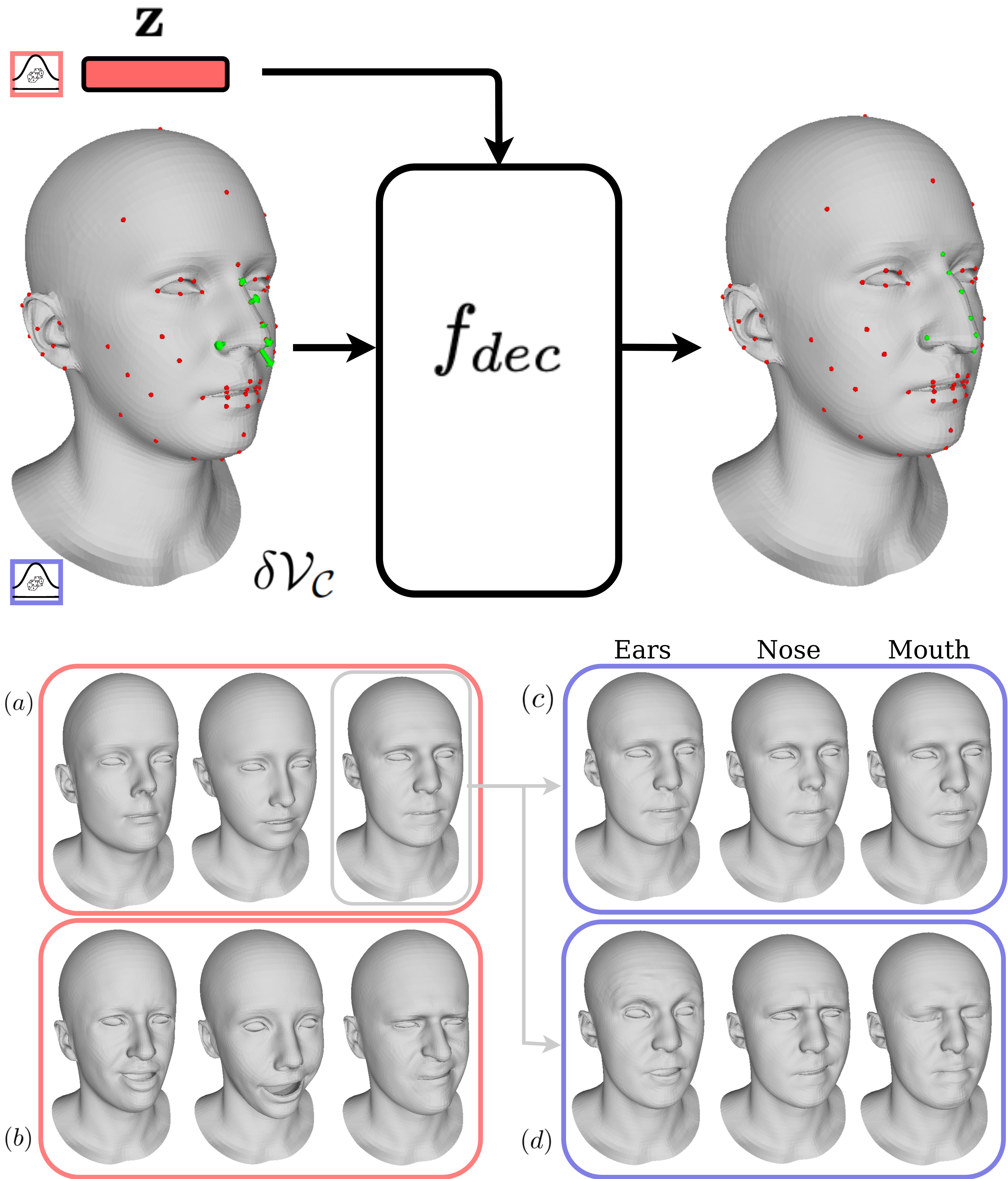} 
    \caption{Overview of the {\it Locally Adaptive Morphable Model (LAMM)} use during inference. (top) Our trained decoder $f_{dec}$ receives as inputs a latent code $\mathbf{z}$ and displacements $\delta \mathcal{V}_{\mathcal{C}}$ over a set of sparse control points (red vertices). Here displacements for control points in the nose region are shown with green arrows. The decoder generates the shape of the object encoded in $\mathbf{z}$, overwriting local geometry to respect $\delta \mathcal{V}_{\mathcal{C}}$. (bottom) We can sample latent codes to generate new instances of human heads in the (a) identity, (b) expression space. Similarly, we can either randomly sample or provide vertex displacements manually at control points to manipulate (c) local identity features (ears, nose, mouth shown here) and (d) add expressions while retaining identity features unchanged.} 
    \label{fig:regions_control} 
\end{figure} 

The capacity to generate and manipulate digital 3D objects lies at the core of a multitude of applications related to the entertainment and media industries \cite{media1,Chai_2023_CVPR}, virtual and augmented reality \cite{avatar1, Lattas_2023_CVPR, Paraperas_2023_ICCV, rodin, stengel2023} and healthcare \cite{medical1, medical2}. In particular, the ability to manually shape virtual humans can result in highly realistic avatars, while granting ultimate creative control to individual users. 

However, achieving fine control in mesh manipulation necessitates the learning of a disentangled representation of 3D shapes which is still an open research problem \cite{Egger20Years, foti1, foti2}. Recently proposed methods based on Graph Convolutional Network (GCN)-based Auto-Encoders (AEs) \cite{coma,spiral,gong2019spiralnet++} have demonstrated impressive performance in dimensionality reduction but typically learn a highly entangled latent space making them unsuitable for detailed shape manipulation. Additionally, despite having a low parameter count, these methods struggle to handle high-resolution meshes, limiting their applicability. Few works \cite{foti1, foti2} have dealt with the disentanglement of local identity attributes, however these methods still rely on GCNs and opt for controlling manipulations through the state of the latent code which is partitioned and assigned to predefined object regions. 
Using the latent code to drive shape manipulation requires the use of explicit optimization objectives to learn a disentangled latent space. Moreover, partitioning its state is critically suboptimal for learning compressed representations of 3D objects. 
We propose a different paradigm which does not involve partitioning the latent code or relying on its state to drive changes in shape, resulting in state-of-the-art (SOTA) disentanglement and reconstruction capabilities in a unified architecture. Instead, we use a global latent code for 3D object unconditional generation and utilise additional inputs to jointly train our generative model to locally overwrite the latent encoded geometry. 
Our main contributions are the following:
\begin{itemize}[leftmargin=4mm] 
    \item We present the {\it Locally Adaptive Morphable Model} (LAMM), a general framework for manipulating the geometry of registered meshes. To the best of our knowledge, this is the first method that allows direct shape control with a single forward pass. Applied on human 3D heads, LAMM exhibits SOTA disentanglement properties and allows for very fine geometric control over both facial identities and expressions.
    \item Our models, trained for manipulation, concurrently exhibit SOTA performance in mesh dimensionality reduction compared against methods trained exclusively on this task. As a result, a single model can be used to generate entirely new shapes and apply both localized and global modifications to their geometry.
    \item We show how our framework can leverage direct control as a primitive to achieve higher level editing operations such as region swapping and sampling. 
    \item By deviating from GCN-based AE design, LAMM can scale to much larger meshes, needs only a fraction of GPU memory to train and can be significantly faster during inference compared to competing methods. 
    For example, trained on 72k vertex meshes with batch size 32, our model requires 7.5Gb of GPU memory and runs at $0.045s$ on a single CPU thread. This model outperforms SpiralNet++ \cite{gong2019spiralnet++} which equivalently requires $>$40Gb of memory and runs $\times 13$ slower at $0.58s$.
    \item We release training and evaluation codes to support future research. Additionally we will release pre-trained models and a Blender add-on \cite{blender} to facilitate intuitive mesh manipulation and fine-grained control.
\end{itemize}

\section{Related Work}\label{sec:related}

{\bf 3D Morphable Models}. Blanz and Vetter pioneered an innovative approach for synthesis and reconstruction of 3D human faces through their seminal {\it 3D Morphable Model} (3DMM) \cite{bfm}. First, they establish a fixed mesh topology, ensuring that each face vertex is semantically significant and could be uniquely identified, e.g. the "tip of the nose". Subsequently, Principal Component Analysis (PCA) is employed to map 3D shapes and textures into a lower dimensional latent code. By adjusting these codes the PCA-based 3DMM can seamlessly generate new facial instances. Such 3DMMs are simple and very robust in terms of performance. They are well established as exceptionally reliable frameworks for 3D face/body analysis, finding applications in numerous human centric models \cite{bfm, lsfm, flame, ploumpis2020towards, ploumpis2019combining, Egger20Years, gecer2019ganfit, Lattas_2023_CVPR,ploumpis20223d} which remain relevant to this date.

However, the linear formulation of PCA limits its expressivity and capacity to capture high frequency details. 
In an effort to capture nonlinear variations in the shape of human faces performing expressions \cite{coma} were the first to utilise Graph Convolutional Networks (GCN) for 3D dimensionality reduction. Their {\it Convolutional Mesh AutoEncoder} (COMA) makes use of fast and efficient spectral convolutions applied on a mesh surface \cite{chebynet} to compress and generate 3D meshes. 
An inductive bias more suitable for registered meshes is proposed in \cite{bouritsas} who incorporate the spiral operator \cite{spiral} in a GCN-based AE. This operator uses a spiral sequence to define an explicit local ordering for aggregating node features, thus giving the convolution kernels a sense of orientation across the mesh surface and greater expressive power. Following a similar principle, \cite{gong2019spiralnet++} further improve the performance of this architecture by enabling context aggregation across larger areas through dilated spiral operators. 
The deep GCN-based family of models discussed above have been consistently shown to outperform PCA at high levels of compression. However, when increasing the size of the latent space to values that are practical for applications, e.g. monocular 3D face reconstruction \cite{gecer2019ganfit,booth20173d}, PCA is shown to match and quickly surpass GCN architectures. We argue that the translation equivariant property of GCNs is not optimal for modelling registered meshes as it does not account for the distinct semantics of each vertex. In contrast, by employing fully connected layers to encode input shapes into tokens and a global receptive field for downstream processing, our framework respects this attribute of the data, having the capacity to treat similar features differently depending on their location.

\begin{figure*}[t]
  \centering
  \includegraphics[width=0.95\linewidth]{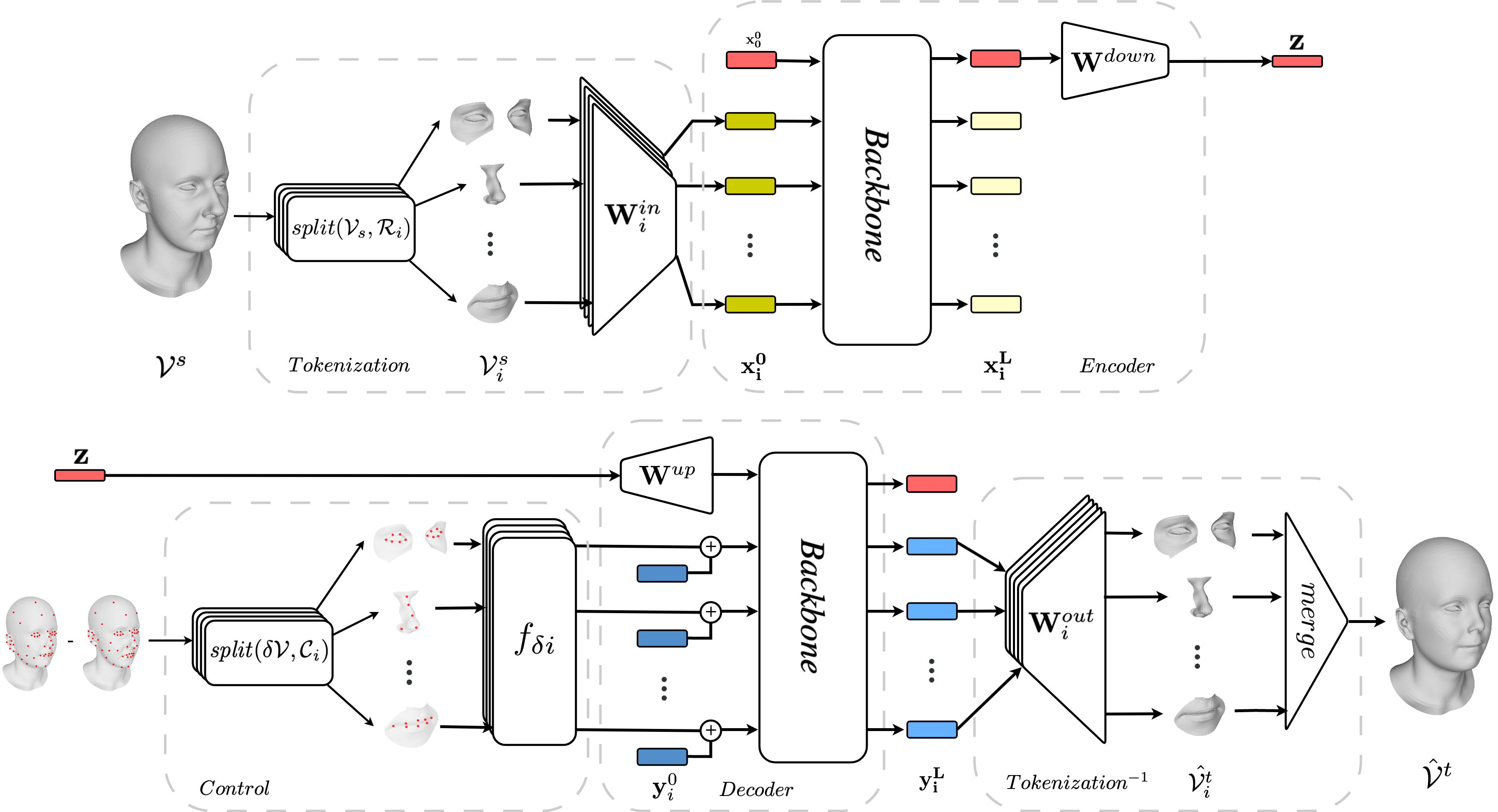}
   \caption{Architecture overview. A {\it source} mesh $\mathcal{V}^s$ is encoded into latent code $\mathbf{z}$ and decoded using additional displacements at control points $\delta\mathcal{V}_C$. (top) Tokenization and encoder modules. A 3D input mesh $\mathcal{V}^s$ is split into regions $\mathcal{V}^s_i = \mathcal{V}^s[\mathcal{R}_i]$, each region is tokenized via region-specific linear weights $\mathbf{W}_i^{in}$ and compressed by the encoder module into latent code $\mathbf{z}$. (bottom) Displacement control, decoder and inverse tokenization modules. User displacements at control points are split into corresponding regions $\mathcal{C}_i$, processed by region specific control networks $f_{\delta i}$ and used by the decoder to overwrite the geometry encoded in $\mathbf{z}$. The inverse tokenization module translates decoder outputs into region geometries, via region-specific linear weights $\mathbf{W}_i^{out}$, which are merged together, using $\mathcal{R}_i$, into the {\it target} estimate $\mathcal{\hat{V}}^t$.}
   \label{fig:method_overview}
\end{figure*}

{\bf Face Manipulation. }
All statistical models of 3D meshes discussed above can be used to manipulate mesh geometry, e.g. the optimal fit for a given shape can be discerned through latent code optimization. The main issue with this approach is that latent spaces learnt for generative modelling are typically highly entangled \cite{Egger20Years, armstrong, foti1}. This attribute discards the possibility for finely-tuned control of local geometry as varying any one dimension of the latent code can concurrently impact multiple properties of the reconstructed data. 
Numerous studies have been proposed to specifically address the challenge of entangled representations of 3D face data for shape manipulation \cite{Egger20Years}. They predominantly utilize face regions to disentangle a latent representation and achieve local control through optimization, either by employing linear models~\cite{Ghafourzadeh2021} or GCN~\cite{Aliari2023, yan2022neo, foti1, foti2}. Notably, works by \cite{foti1, foti2} adeptly deform 3D face geometry to manipulate an individual's identity. Both works harness the backbone architecture from~\cite{gong2019spiralnet++}, along with various face regions, each governed by non-overlapping parts of the latent code. In \cite{foti1} their AE learns a disentangled latent space by creating a composite batch through swapping arbitrary features across different samples. This enables the definition of a contrastive loss that leverages known differences and similarities in the latent representations. In a related approach, \cite{foti2} also segment their latent codes to region specific parts and use a loss component forcing segments to adhere to the local eigenprojections of identity attributes.
Similar to these works, we also employ predefined regions to segment our inputs. Importantly, we do not partition our latent codes into region-specific segments, which is beneficial for retaining strong performance in mesh reconstruction. Furthermore, we do not enforce spatial disentanglement through explicitly defined loss components (we train only with $L_1$ loss) but rather reach SOTA disentanglement performance solely as an emerging property of our architecture.

\section{Method}\label{sec:method}
In this section we present the LAMM framework for the disentangled direct manipulation of 3D shapes. First, we describe the architecture, followed by a self-supervised training scheme for learning direct local control over mesh geometry.
In the following, we assume a 3D mesh is represented as $\mathcal{M}=\{\mathcal{V},\mathcal{F}\}$, where $\mathcal{V} \in \mathbb{R}^{N \times 3}$ is a set of $N$ vertices represented as points in 3D space, and $\mathcal{F}$ is a set of faces defining a shared topology. 
\subsection{Architecture}
We leverage the AE framework for mesh representation learning, with targeted modifications in the decoder architecture to facilitate direct manipulation of mesh geometry.
First, a {\it source} mesh $\mathcal{V}^s$ is encoded into a latent space $\mathbf{z}=f_{enc}(\mathcal{V}^s) \in \mathbb{R}^{D}$ and subsequently decoded into an estimated {\it target} geometry $\mathcal{\hat{V}}^t=f_{dec}(\mathbf{z}, \delta\mathcal{V}_{C})$ using control vertex displacements $\delta\mathcal{V}_{C}$ as additional input to the decoder.
Both encoder and decoder modules utilise patch-based backbones, namely the Transformer \cite{aiayn} and MLPMixer \cite{mlpmixer}, that operate on unordered sets of tokens and exhibit a homogeneous feature space, i.e. feature dimensions stay constant throughout the network. An overview of the proposed architecture is presented schematically in Fig.\ref{fig:method_overview}.

{\bf Tokenization.} For deriving the inputs to the encoder we first split each {\it source} object into $K$ non-overlapping dense regions, $\mathcal{V}^s_i = \mathcal{V}^s[\mathcal{R}_i] \in \mathbb{R}^{N_i \times 3}$, defined by a set of indices, $\mathcal{R}_i = \{j \in \mathbb{N}, 1 \leq j \leq N\}, i=1,...,K$ (see Fig.\ref{fig:regions_control}). 
We further flatten each region into a vector of size $\mathbf{v}_i^s \in 3N_i$ and tokenize it as a linear projection $\mathbf{W}_i^{in} \mathbf{v}_i^s$, where $\mathbf{W}_i^{in} \in \mathbb{R}^{D \times 3N_i}$. 
The proposed tokenization scheme differs significantly from the one used in \cite{vit, mlpmixer} that operate on fixed-size square patches. First, no parameter sharing takes place during this step. Instead there exists a one-to-one mapping between tokenization weights $\mathbf{W}_i^{in}$ and input vertex coordinates that respects the semantic meaning of each vertex. This allows us to define regions with varying number of vertices that are designed to include distinct object parts based on the template mesh, e.g. a "whole eye" in human face meshes. 
Finally, our regions are typically much larger than commonly employed in computer vision \cite{vit, mlpmixer, vitsforsits}, which results in a small number of tokens. For example, while the original implementation of ViT \cite{vit} used 256 tokens for image classification, we use as few as 11 tokens for 3D face modelling. Since the Transformer space and time complexities are both quadratic to the number of tokens, our models are very efficient in terms of memory consumption, can scale to very large meshes and can run significantly faster than GCNs especially on a CPU. 

{\bf Encoder.} Our fixed size tokenized representation for region $i$ is $\mathbf{x}_i^0 = \mathbf{W}_i^{in} \mathbf{v}_i^s$. 
To these features we prepend a learnable identity token $\mathbf{x}_{0}^0 \in \mathbb{R}^{D}$ \cite{bert,deit} which is independent of the input. After processing by the $L$-layer homogeneous encoder, we retain only the state of the first index token $\mathbf{x}_{0}^L \in \mathbb{R}^{D}$ that forms the encoded representation of the {\it source} geometry. To obtain our latent code we linearly project this vector into a space of desired dimensionality $\mathbf{z} = \mathbf{W}^{down} \mathbf{x}_{0}^L \in \mathbb{R}^{d}$. The tokenization and encoder modules are illustrated in the top part of Fig.\ref{fig:method_overview}.

{\bf Decoder.} In the decoder module, we first linearly project the latent code back into a $D$ dimensional vector $\mathbf{y}_0^0 = \mathbf{W}^{up} \mathbf{z}$. 
Analogously to the encoder, we build our decoder inputs by appending $K$ additional learned tokens $\mathbf{y}_i^{0} \in \mathbb{R}^D$ to the projected latent code $\mathbf{y}_0^0$ \cite{detr,vitsforsits}.
We then select a set of non-overlapping, sparse vertex indices $\mathcal{C}_i \subseteq \mathcal{R}_i$ that are used as control points (Fig.\ref{fig:regions_control}). 
We gather all desired displacements at control points per region $\delta\mathcal{V}_{Ci} = (\mathcal{V}^t - \mathcal{V}^s)[Ci] \in \mathbb{R}^{3|\mathcal{C}_i|}$ and process each with a fully connected network $f_{\delta i}$ the output of which is added to respective learned tokens $\mathbf{y}_i^{0}, i \geq 1$.
To naturally make our method work as an AE, we design $f_{\delta i}$ without any bias terms. In this manner, for the special case of $\delta\mathcal{V}_{Ci} = 0$ we get $f_{\delta i}(0) = 0$ and LAMM reproduces the behaviour of an AE. 

{\bf Tokenization}$\mathbf{^{-1}}$. Following the inverse process of region splitting and tokenization steps we now project each token independently to a vector $\hat{\mathbf{v}}_i^t = \mathbf{W}_i^{out} \mathbf{y}_{i}^L  \in \mathbb{R}^{3N_i}$, reshape its dimensions to $N_i \times 3$ and use $\mathcal{R}_i$ to merge regions into a unified 3D template mesh, $\hat{\mathcal{V}^t}$. The decoder and inverse tokenization modules are presented schematically in the bottom part of Fig.\ref{fig:method_overview}.

 \begin{figure}[t]
    \centering 
    \includegraphics[width=0.375\textwidth]{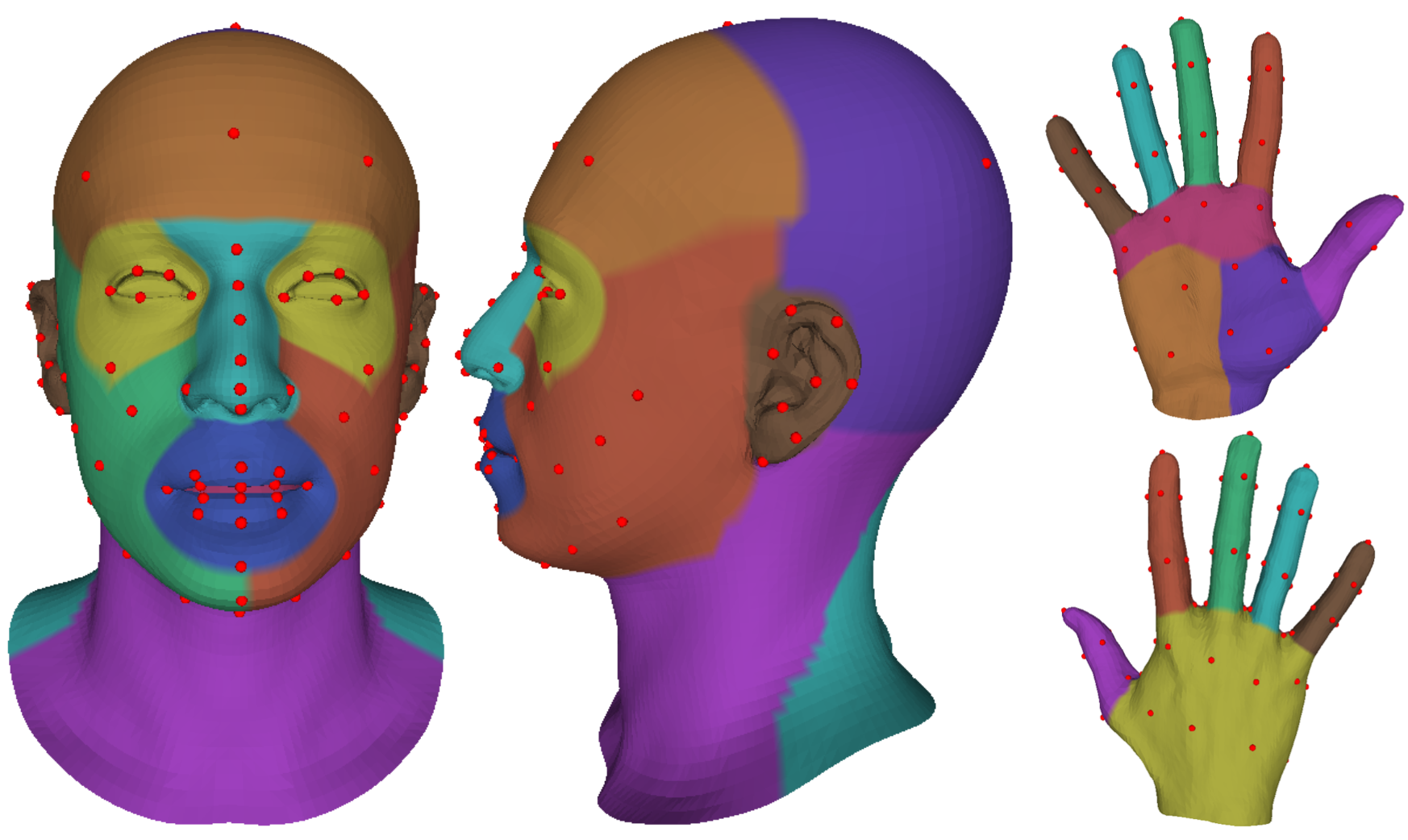} 
    \caption{Templates for UHM12k and Handy data. Colors indicate dense regions $\mathcal{R}_i$. Control points $\mathcal{C}_i$ are shown in red dots.} 
    \label{fig:regions_control} 
\end{figure} 

\subsection{Training}\label{sec:training}
{\bf Self-supervised objective.}
At each step we sample $B$ {\it source} and {\it target} elements $\mathbf{V^s}, \mathbf{V^t} \in \mathbb{R}^{B \times N \times 3}$ from the training set and construct our {\it source} and {\it target} batches respectively as $\{\mathbf{V}^s, \mathbf{V}^s\}$ and $\{\mathbf{V}^s, \mathbf{V}^t\}$. In this manner, the first half of our batch is used for AE training and the second half for learning {\it source-to-target} transformations. For identity manipulation we use a random permutation of $\mathbf{V}^s$ as $\mathbf{V}^t$. For expression manipulation, $\mathbf{V}^s$ and $\mathbf{V}^t$ include the same identities with neutral and non-neutral expressions.

Training with direct supervision from real {\it target} data has the benefit of providing realistic examples for the {\it target} geometry space. However, there are a few issues with this approach. In practice, interactive vertex control is an iterative process. As a result typical user input displacements are expected to be smaller and significantly sparser than the ones that globally transform one sample into another. To reduce the effect of displacement size discrepancy we train with a linear combination of {\it source} and {\it target} samples $\alpha \mathcal{V}^{s} + (1 - \alpha) \mathcal{V}^{t}, \alpha \sim \mathcal{U}(\alpha_{min}, 1)  $ as the new {\it target} values.
This effectively reduces the absolute values of displacements encountered during training in addition to augmenting the {\it target} space. 
The sparsity discrepancy is a trickier problem as it is not trivial to find realistic training pairs with only localized differences. 
We find that this issue is solved to a satisfactory degree implicitly as the proposed architecture naturally propagates deformations only to non-zero-displacement regions. We believe this attribute to be an intrinsic property of our architecture. In support of this claim, we show in the supplementary material that only through AE training, i.e. $\delta\mathcal{V}_{Ci} = 0$, corrupting the values of learnable decoder tokens $\mathbf{y}_i^0, i>0$ influences only their corresponding region geometries. 

{\bf Multilayer Loss.} To adequately train our architecture, a distance-based loss between the {\it target} and output geometries is applied which suits our fundamental learning objective. In all experiments we utilize the $L1$ distance metric, calculated as $||\hat{\mathcal{V}^t} - \mathcal{V}^t||_1$. We introduce an extension to this loss, applied at every level of the model, which increases the encoder's capacity to encode input-related information and guides the decoder transformation from {\it mean} to {\it target} shape. 
Both employed backbones \cite{aiayn, mlpmixer} produce homogeneous feature spaces of size $D$ across all encoder and decoder layers. 
At every layer $l$ we decode region features $\mathbf{x}_i^l, \mathbf{y}_i^l, i \geq 1$ using $\mathbf{W}_i^{out}$ and merge regions together to obtain a multilayer output geometry.  

The objective of the encoder is to obtain a compressed representation of the input in the form of the latent code $\mathbf{z}$. At input, we want our tokens to retain as much information about the input shape to begin with, so we add a loss component $||\mathbf{W}_i^{out} \mathbf{x}_i^0 - \mathbf{v}_i^s||_1$ that encourages the input tokens to decode close to the {\it source} geometry. Furthermore, we provide some guidance to our encoder to progressively discard input related information from the region features through a loss component $||\mathbf{W}_i^{out} \mathbf{x}_i^L - \bar{\mathbf{v}}_i||_1$ that encourages the final encoder region tokens to match the mean shape for each region $\bar{\mathbf{v}}_i$ over the training data. 

Similarly, we expand this pattern for all intermediate layers in a linear fashion.

\begin{equation}\label{eq:enc_loss}
    L_{enc} = \sum_{l=0}^L \frac{\lambda^l}{K} \sum_{i=1}^K || \mathbf{W}_i^{out} \mathbf{x}_i^l - \frac{1}{L} ((L - l) \mathbf{v}_i^s + l \bar{\mathbf{v}_i}) ||_1 
\end{equation}

In the decoder, we first center initialised tokens $\mathbf{y}_i^0$ through the term $||\mathbf{W}_i^{out} \mathbf{y}_i^0 - \bar{\mathbf{v}_i}||_1$ which encourages them to decode into the mean shape. The output of the final layer will be our model's estimate of the object geometry, thus, we supervise through the term $||\mathbf{W}_i^{out} \mathbf{y}_i^L - \mathbf{v}_i^t||_1$. Similarly, we apply a multilayer loss to intermediate layers to encourage a progressive transformation of the region tokens from {\it mean} to {\it target} geometry. 

\begin{equation}\label{eq:dec_loss}
    L_{dec} = \sum_{l=0}^L \frac{\lambda^l}{K} \sum_{i=1}^K || \mathbf{W}_i^{out} \mathbf{y}_i^l - (\frac{1}{L} (l \mathbf{v}_i^t + (L - l) \bar{\mathbf{v}_i}||_1
\end{equation}

In both eqs.\ref{eq:enc_loss}, \ref{eq:dec_loss} $\lambda_l$ is a weight for the loss component at layer $l$. In practice we opt for a simple solution of keeping all $\lambda_l = 1$ which leads to stable training and good generalization performance. 
We note that in both equations the inverse tokenization weights $\mathbf{W}_i^{out}$ are shared in every loss term, forcing a common geometric representation throughout the network which progressively transforms from {\it source} to {\it mean} to {\it target}. Additionally, our data are typically mean centered prior to any processing, thus, $\bar{\mathbf{v}_i}$ reduces to the zero vector. In this case, our multilayer loss can be viewed as a form of regularization forcing the values of $\mathbf{W}_i^{out}$ to be either small in values or orthogonal to the learnt tokens $\mathbf{y}_i^0$.

\begin{figure*}[t]
  \centering
  \includegraphics[width=0.96\textwidth]{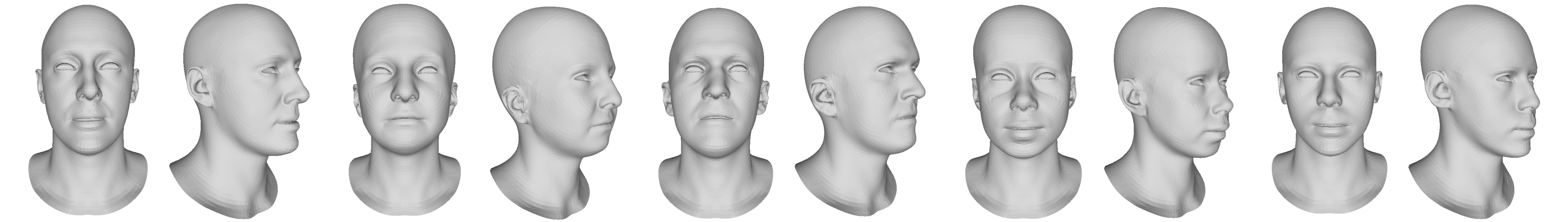} 
    \includegraphics[width=0.96\textwidth]{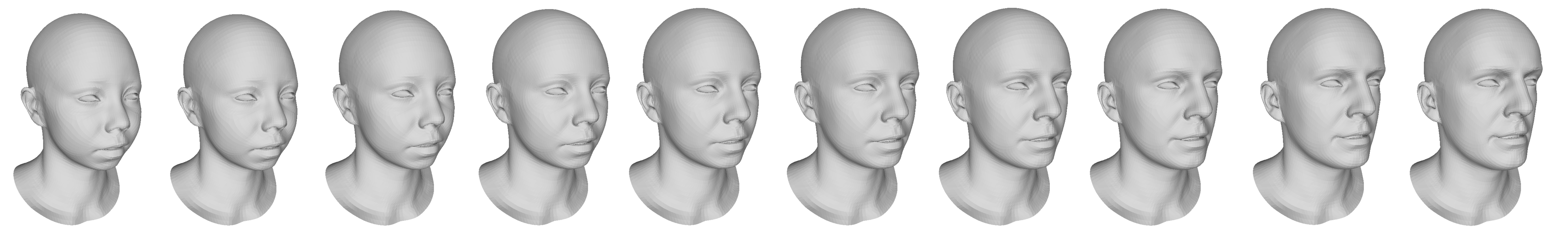} 
    \caption{Qualitative assessment of trained AE in UHM12k. (top) New shape generation through sampling of the latent space $\mathbf{z}$, (bottom) Interpolation between latent codes for two samples from the evaluation data shows a smooth transition between identities.} 
    \label{fig:compression_generation_interpolation} 
\end{figure*}

\section{Experiments}\label{sec:experiments}
\subsection{Implementation details}
{\bf Datasets.} For effectively learning disentangled human head identity manipulations we need a large scale dataset of 3D heads in neutral expression. For this reason, we register 6k heads from the original data used in the UHM model \cite{ploumpis2020towards} into a new quads-based template which consists of 12k vertices, is more friendly towards rigging and is animation ready. 
We will refer to this data as UHM12k. To create our synthetic expression data we applied a set of 28 blendshapes \cite{3dscanstore} to the UHM12k data. 
To test our method on high resolution meshes, we utilise the linear models provided by the UHM \cite{ploumpis2020towards} and sample 10k identities consisting of 72k vertices. 
To evaluate our method on hand data, we utilized Handy \cite{potamias2023handy}, a large-scale hand model, trained from more than 1k distinct identities. In particular, to generate our hand dataset, we randomly sampled 10k hands from the Handy latent space. All datasets were split into training and evaluation sets at a 9-1 ratio. 

\begin{figure*}[t]
    \centering 
    \includegraphics[width=0.93\textwidth]{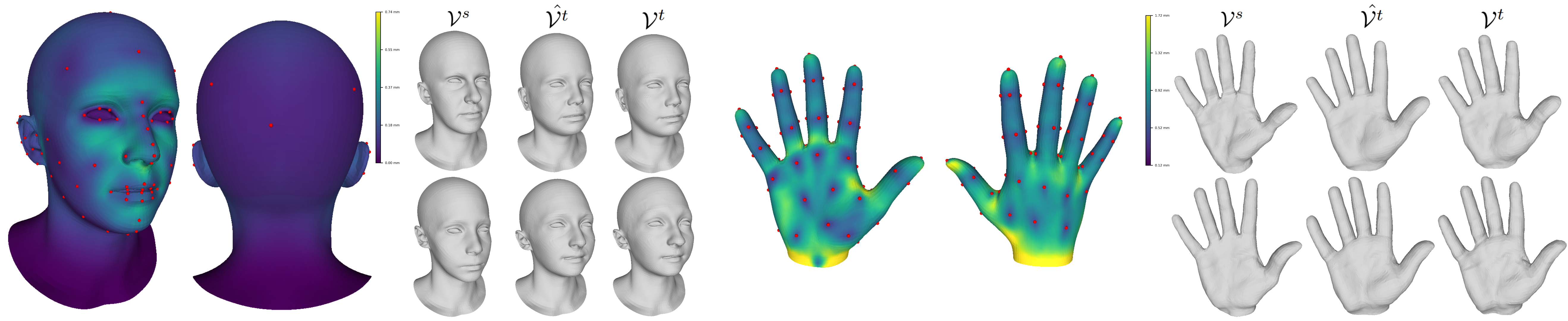} 
    \caption{Identity manipulation performance under direct vertex control for UHM12k (left) and Handy (right) data. We observe that errors are typically reduced close to control points. For both datasets our method learns to produce highly realistic output geometries.} 
    \label{fig:identity_control} 
\end{figure*}

\begin{figure*}[t]
    \centering 
    \includegraphics[width=0.93\textwidth]{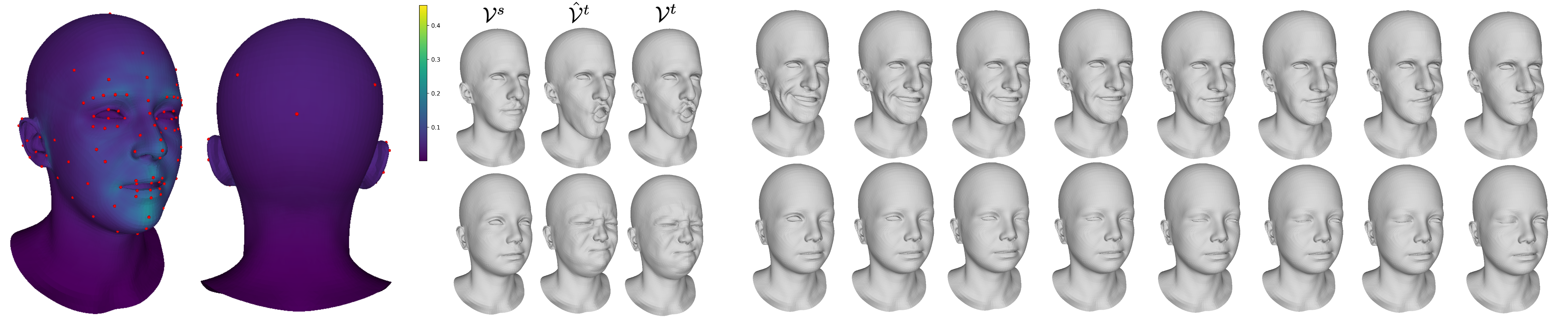} 
    \caption{Expression manipulation performance under direct vertex control for UHM12k data. (left) We observe that per-vertex errors are very small and that $\hat{\mathcal{V}}^t$ closely resembles the appearance of $\mathcal{V}^t$. (right) Linear interpolation in expression intensities for the two samples. We observe a smooth transition between facial expressions and the ability to perform highly localized edits by manipulating eyes independently.} 
    \label{fig:expression_control} 
\end{figure*}

{\bf Architecture.} We optimize our architecture through an ablation study on autoencoding using the UHM12k data, presented in the supplementary material. Overall, our architecture consists of five encoder and three decoder layers with 512 feature dimension. For manipulation experiments we split the UHM12k and the Handy templates into 11 and 9 regions and utilise 73 (101 for expression modelling) and 72 control points respectively, all of which are illustrated in Figs.\ref{fig:regions_control}, \ref{fig:identity_control}, \ref{fig:expression_control}. For identity control with the UHM data we use the same face regions as \cite{foti1, foti2} and 91 control points.

{\bf Training. }
Our proposed architecture can train for significantly longer than GCN baselines without observing a drop in evaluation set performance. We train all networks (including baselines) in mesh reconstruction and manipulation for 1,500 epochs on a single Nvidia Titan X GPU, with batch size 32 (adjusted for GCNs according to memory). We use the AdamW optimizer \cite{adamw} starting with a 10 epoch learning rate warm up to a $10^{-4}$ which is decayed via a 1-cycle cosine decay \cite{loshchilov2017sgdr} to $10^{-6}$. For local control experiments, we find it is crucial to initialize our models from checkpoints pre-trained exclusively for autoencoding. Further details are provided in the supplementary material.

\subsection{3D shape reconstruction}\label{sec:results_compression}
In Table \ref{tab:results_dim_reduction} we compare the reconstruction performance of our proposed framework for autoencoding against PCA \cite{blanz_vetter} and state of the art GCNs \cite{coma,bouritsas,gong2019spiralnet++}. Overall, our method is shown to significantly outperform others applied in head and hand data with improvements being more pronounced on UHM12k. 

Comparing the two employed backbones, the MLPMixer is a consistently strong performer, surpassing baselines in all cases tested. Transformer-based backbones are found to be more suitable for the expression enhanced data, but weaker than the MLPMixer for the UHM and Handy data.
In Fig.\ref{fig:compression_generation_interpolation} (top) we present generated samples drawn from our head model. Only the decoder part of our architecture is used here. Similar to previous works \cite{blanz_vetter, coma, bouritsas, gong2019spiralnet++} sampling new instances of global shape can be achieved by fitting a Gaussian probability distribution to $\mathbf{z}$ values collected over the training data. 

In the bottom part of Fig.\ref{fig:compression_generation_interpolation} we choose two highly diverse identities from the evaluation set, encode both and present geometries generated by linear interpolation of their latent codes, noting a smooth transition between generated shapes.

\begin{table}[t]
    \centering
    \footnotesize
    \caption{Quantitative evaluation of 3D shape reconstruction for models trained exclusively in autoencoding with latent size 256. Presented values are mean Euclidean distances ($\times10^{-2}$mm).}
    \begin{tabular}{l|cc|c|c}
     & UHM12k & +expr.& UHM & Handy \\
    \hline 
    \hline
     PCA & 10.42 & 11.49 & 11.73 & 25.90 \\
     COMA \cite{coma}  & 13.11 & 14.20 & 15.40 & 27.20\\
     Neural3DMM \cite{bouritsas} & 11.82 & 12.97 & 13.88 & 26.75\\
     SpilarNet++ \cite{gong2019spiralnet++} & 11.53 & 12.58& 13.55 & 26.72\\
     \hline
     LAMM-Transformer &  9.08  & {\bf 9.09} &  13.70 & 26.31 \\
     LAMM-MLPMixer &  {\bf 7.97} & 9.51 & {\bf 11.48} & {\bf 24.60}\\
    \hline
    \end{tabular}
    \label{tab:results_dim_reduction}
\end{table}

\subsection{Local control}
{\bf Global manipulation. } We present a quantitative evaluation of models trained in shape manipulation in Table \ref{tab:manipulation_compression}. First, we note that AE performance is very similar to the values presented in Table \ref{tab:results_dim_reduction} where models are trained exclusively in mesh reconstruction. In fact, AE performance is found to improve in many cases suggesting the complementarity of the two tasks. We also  find that expression training can increase AE performance significantly. If we train for expression manipulation but select our model based on its AE performance our Transformer based model achieves a distance error of $7.79 \times 10^{-2}$mm, which is $14\%$ less compared to the best value we obtained from AE training in Table \ref{tab:results_dim_reduction} ($9.08 \times 10^{-2}$mm).  
Figs.\ref{fig:identity_control}, \ref{fig:expression_control} illustrate per-vertex mean prediction errors plotted over respective templates for identity and expression manipulations. In Fig.\ref{fig:identity_control} these are shown to be clearly smaller close to control points verifying the metrics reported in Table \ref{tab:manipulation_compression}.
For human heads, errors are mostly concentrated in the frontal region of the face. For hands, errors are significantly larger and more evenly spread. However, there are clear concentrations near the wrist which can be attributed to the absence of control points in these regions. We additionally show examples of {\it source} to {\it target} generations from the evaluation data, observing how predicted geometries match the appearance of {\it target} for both datasets.
For expression manipulation, we can generally observe from Fig.\ref{fig:expression_control} and Table \ref{tab:manipulation_compression} that errors are smaller compared to identity manipulation. As expected, we observe that errors are also concentrated in the face region where most of the deformation takes place and particularly small in head regions that do not participate in facial expressions such as the ears and skull. We also note that, in contrast to identity manipulation, errors are significantly lower here at non-control points which can be attributed to the fact that $\notin \mathcal{C}_i$ includes many vertices with very small deformations, e.g. skull. \\

\begin{figure}[h]
    \centering 
    \includegraphics[width=0.45\textwidth]{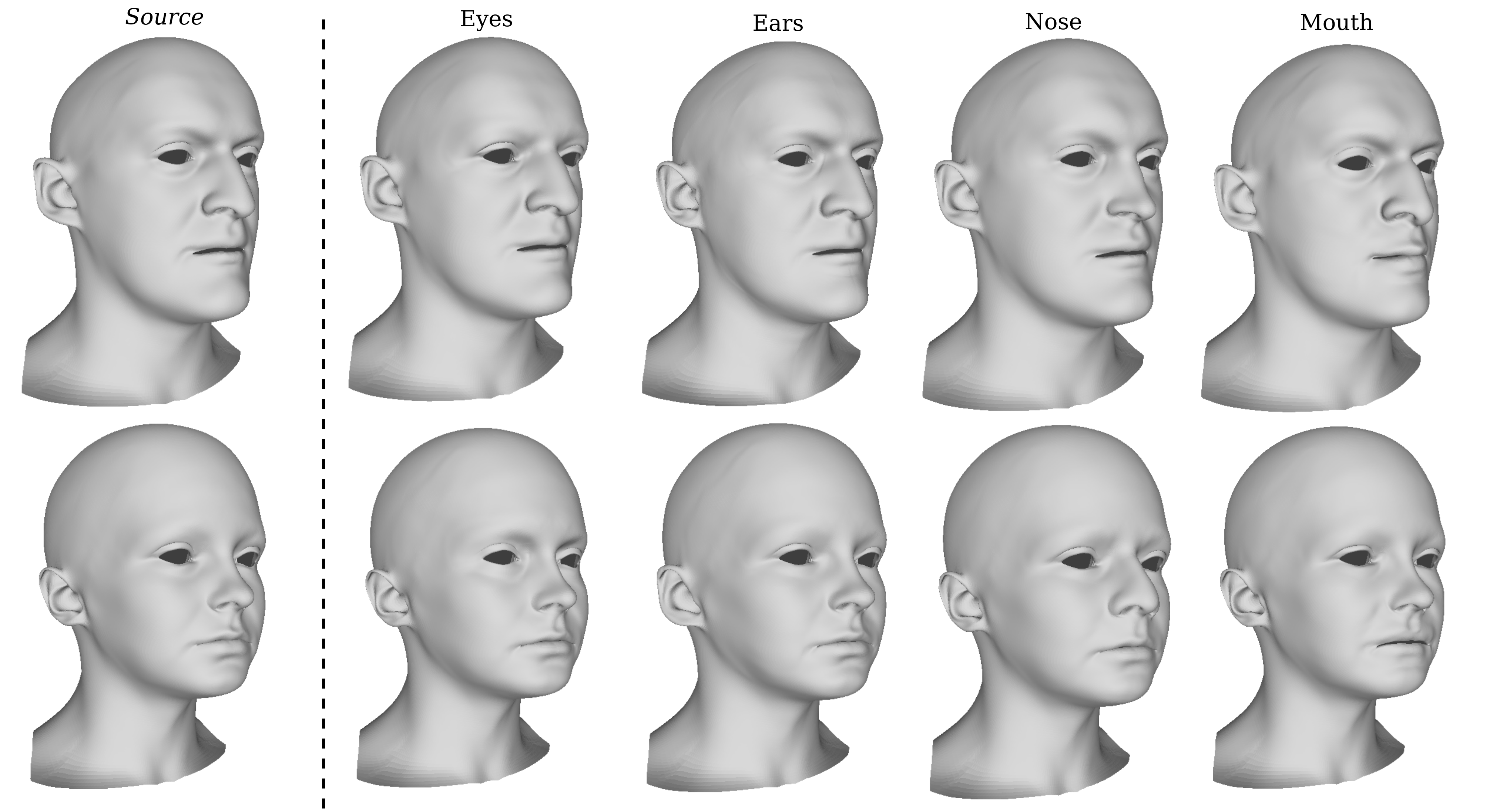} 
    \caption{Head region swapping for UHM data.  } 
    \label{fig:region_swapping} 
\end{figure}

\begin{table}[t]
    \centering
    \footnotesize
    \caption{Quantitative evaluation of models trained in 3D shape manipulation. We present mean Euclidean distance error ($\times10^{-2}$mm) for the AE case ($\mathcal{V}^t = \mathcal{V}^s$), control ($\in C$) and remaining ($\notin C$) vertices. We note AE performance can surpass that of models trained exclusively for mesh reconstruction.
    $\dagger$ neutral expression only.}
    \begin{tabular}{c|ccc|ccc}
     \multicolumn{1}{c}{} & \multicolumn{3}{c}{UHM12k} & \multicolumn{3}{c}{UHM12k+expr.} \\
     Backbone & AE & $\in C$ & $\notin C$ & AE$\dagger$ & $\in C$ & $\notin C$\\
    \hline 
     Transformer & 8.79 & {\bf 18.47} & {\bf 26.03} & {\bf 8.61} & {\bf 15.86} & {\bf 9.67} \\
     MLPMixer & {\bf 8.36} & 19.87 & 28.08 & 8.98 & 15.92 & 9.73 \\
    \hline

     \multicolumn{1}{c}{} & \multicolumn{3}{c}{UHM} & \multicolumn{3}{c}{Handy} \\
     Backbone & AE & $\in C$ & $\notin C$ & AE & $\in C$ & $\notin C$\\
    \hline 
     Transformer & 15.86 & 26.23& 38.80 &  26.78& 28.32 & 55.69 \\
     MLPMixer & {\bf 11.63} & {\bf 17.97} & {\bf 32.03} & {\bf 24.59} & {\bf 25.57} & {\bf 51.66} \\
    \hline
    
    \end{tabular}
    \label{tab:manipulation_compression}
\end{table}

\begin{figure*}[h]
  \centering
  \includegraphics[width=0.94\textwidth]{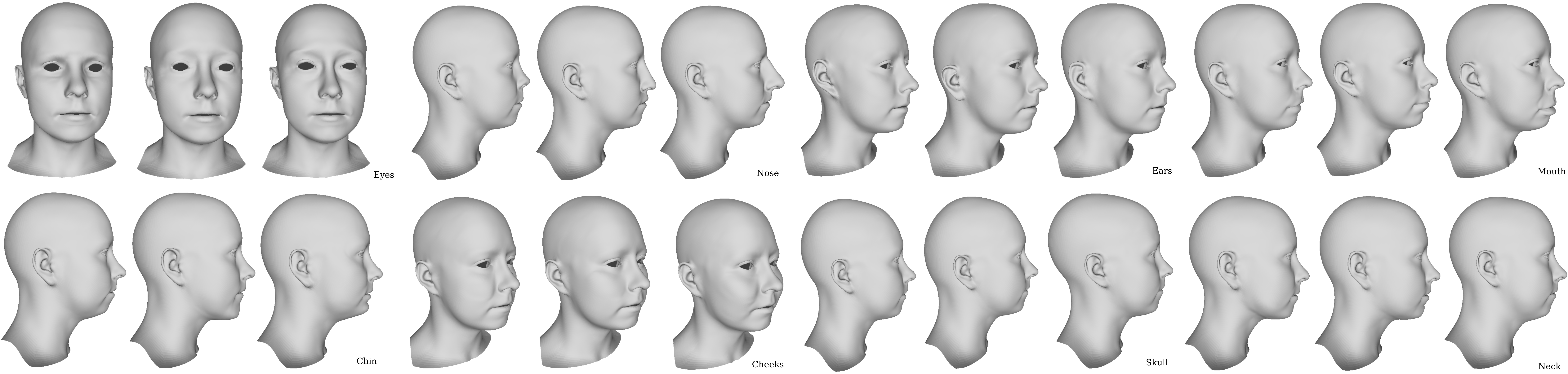} 
   \vspace{0.5cm}
    \includegraphics[width=0.94\textwidth]{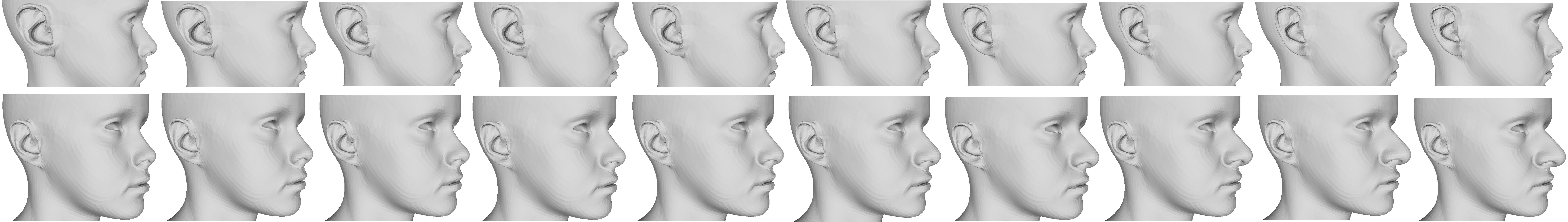} 
    \caption{Random generation and interpolation of head regions.(top) Random head region generation on UHM (bottom) Region interpolation for ears and nose using UHM12k. We observe a smooth transition between generated regions while remaining areas are unaffected.} 
    \label{fig:region_sampling} 
\end{figure*}

\begin{figure*}[t]
    \centering 
    \includegraphics[width=0.94\textwidth]{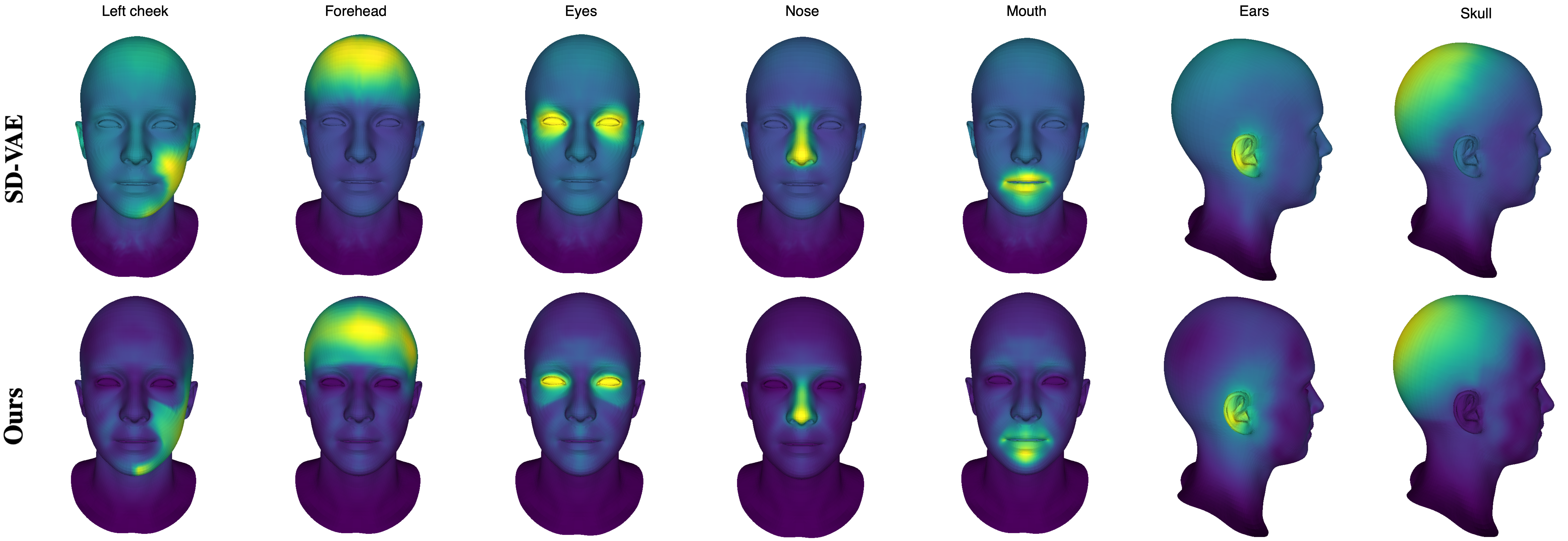} 
    \caption{Evaluation of disentanglement performance compared to state-of-the-art SD-VAE \cite{foti1}. For all data in the evaluation set we randomly sample 20 shapes for each head region. Plotted values are per-vertex mean Euclidean distances between input and output shapes.} 
    \label{fig:exp_disentanglement} 
\end{figure*}

\vspace{-0.45cm}
{\bf Region swap. } For swaping regions among meshes it is critical to find a set of control point displacements that best transform the {\it source} region $\mathcal{V}_i^{s}$ to a desired {\it target} shape $\mathcal{V}_i^{t}$. However, calculating displacements directly results in non realistic geometries as the two regions may be offset by a significant amount. To resolve this issue, we first rigidly align regions by their common outer edge indices.%
Due to the simplicity of applied transformations, rigid alignment is found not to disturb the appearance of the {\it target} shape, keeping it realistic. The outcome of this operation is reduced offsets at the boundaries, however, there are still discontinuities due to the restricted nature of rigid alignment. It is critical to note, however, that the region's control vertices, sparsely located in its interior, represent a valid sparse geometry since there exists a valid dense shape that fits these vertices while respecting the region's boundary conditions. After finding the optimal rigid transformation, it can be applied to the control points of the {\it target} region to obtain the updated {\it target} locations of the control vertices, which in turn are used to calculate $\delta\mathcal{V}_{Ci}$.
We employ this methodology for region swapping in human heads. Fig.\ref{fig:region_swapping} demonstrates swapping the eyes, ears, nose and mouth regions between two diverse meshes from the UHM data. 

{\bf Region sampling and disentanglement.} We show it is possible to generate new dense region shapes by sampling valid displacements for the control vertices. Using the same approach described for region swapping, we randomly select data pairs from the training data and collect a set of valid control vertex displacements for every region. To these, we fit a Gaussian distribution from which we can sample from to produce new valid control point displacements. We present samples of generated face regions in the top part of Fig.\ref{fig:region_sampling} using the UHM data. In the bottom part, we utilise UHM12k to illustrate face part interpolation for the ears and nose regions.
We evaluate the capacity of our model to have only localized effects in manipulated geometry in Fig.\ref{fig:exp_disentanglement} compared to SOTA method SD-VAE \cite{foti1}.
Consistently, we find that our method leads to more localized effects for all face regions and improved disentanglement over generated geometries.
Importantly, we note that in contrast to \cite{foti1}, our framework is trained without any loss component aimed towards disentanglement which indicates this to be a intrinsic property of our method. We provide additional evidence to support this claim in the supplementary material. Furthermore, we find that SD-VAE has low reconstruction performance at $18.90 \times 10^{-2}mm$, more than double the reconstruction error of our method from Table \ref{tab:manipulation_compression} and more than SpiralNet++ \cite{gong2019spiralnet++} which is the backbone employed by SD-VAE. As we discuss in the supplementary material, this is mainly attributed to the choice of partitioning the latent code made by SD-VAE to control region geometries. 

\section{Conclusion}
We have introduced LAMM, a comprehensive framework for learning 3D mesh representations and achieving spatially disentangled shape manipulation. 
To our knowledge, LAMM is the first end-to-end trainable method that is capable of directly 
manipulating mesh shape, by using desired displacements as inputs, in a single forward pass. 
Our extensive experiments with 3D human head and hand mesh data demonstrate that LAMM simultaneously attains state-of-the-art performance in both disentanglement and reconstruction within a unified architecture. Notably, it scales more effectively to high-resolution data, requires less memory for training, and offers faster inference speeds compared to existing methods.
When implemented within the identity space of human heads, our method exhibits significant potential for applications in face sculpting. Additionally, when trained in the expression space of human heads, LAMM demonstrates considerable promise for manually enhancing digital characters with expressive details and for its application in the field of animation.
Fast CPU inference makes LAMM an energy efficient and economically viable option, democratizing access to advanced 3D modeling.

{
    \small
    \bibliographystyle{ieeenat_fullname}
    \bibliography{ms}
}


\end{document}